\documentclass[letterpaper, 10 pt, conference]{ieeeconf}  

\usepackage{hyperref}
\usepackage{amsmath}
\usepackage{amsfonts}
\usepackage{wrapfig}
\usepackage{xcolor}
\usepackage{graphics} 
\usepackage{epsfig} 
\usepackage{algorithm2e}
\RestyleAlgo{ruled}

\hypersetup{
  colorlinks   = true, 
  urlcolor     = red, 
  linkcolor    = blue, 
  citecolor   = blue 
}

\IEEEoverridecommandlockouts                              

        \title{\LARGE \bf
Convolutional Occupancy Models \\ for Dense Packing of Complex, Novel Objects
}

\author{Nikhil Mishra$^{*, 1, 2}$, Pieter Abbeel$^{1, 2}$, Xi Chen$^{1}$, Maximilian Sieb$^{3}$
\thanks{$^*$Correspondence to: nmishra@berkeley.edu.}%
\thanks{
$^{1}$Covariant.ai, 
$^{2}$UC Berkeley, 
$^{3}$Work done at Covariant.ai}%
}

\begin{document}

\maketitle
\thispagestyle{empty}
\pagestyle{empty}

\begin{abstract}

Dense packing in pick-and-place systems is an important feature in many warehouse and logistics applications.
Prior work in this space has largely focused on planning algorithms in simulation, but real-world packing performance is often bottlenecked by the difficulty of perceiving 3D object geometry in highly occluded, partially observed scenes.
In this work, we present a fully-convolutional shape completion model, F-CON, which can be easily combined with off-the-shelf planning methods for dense packing in the real world.
We also release a simulated dataset, COB-3D-v2, that can be used to train shape completion models for real-word robotics applications, and use it to demonstrate that F-CON outperforms other state-of-the-art shape completion methods.
Finally, we equip a real-world pick-and-place system with F-CON, and demonstrate dense packing of complex, unseen objects in cluttered scenes.
Across multiple planning methods, F-CON enables substantially better dense packing than other shape completion methods. 
\end{abstract}

\section{INTRODUCTION}

Recent years have seen huge commercial interest in robotic pick-and-place systems for applications in warehouse automation and logistics.
While current work on these systems has mostly focused on picking, intelligent placing is also critical to many use cases.
For example, in order fulfillment, a robot must densely pack objects into shipping boxes that will be sent from a warehouse to a customer.
Suboptimal packing performance leads to inefficiencies in the overall operation, as larger boxes or more shipments will be unnecessarily required, increasing both waste and cost.
As a result, \textit{dense packing} -- a task where a robot must pack objects into a given container in a way that maximizes the density or number of objects -- is a requisite feature in many real-world pick-and-place applications.

The majority of work in dense packing has focused on the sequential-decision-making aspect of the problem: in order to achieve optimal packing densities, every object needs to be placed carefully, with regard for how it affects the subsequent objects that will need to be packed into the same container.
The result of this line of work has been a series of attempts to cast dense packing as a reinforcement learning (RL) problem \cite{rl1} \cite{rl2} \cite{rl3}.
To make this difficult problem more tractable and easier to evaluate, common practices have been to operate in simulation on simplified state representations, such as by approximating all objects as cuboids, or to assume that complete state information is available, such the ground-truth geometry of all objects \cite{hm} \cite{real1}.
However, there remain perceptual challenges that need to be addressed in order to apply this work to the real world.

\begin{figure}[thpb]
  \centering
  \includegraphics[width=0.48\textwidth]{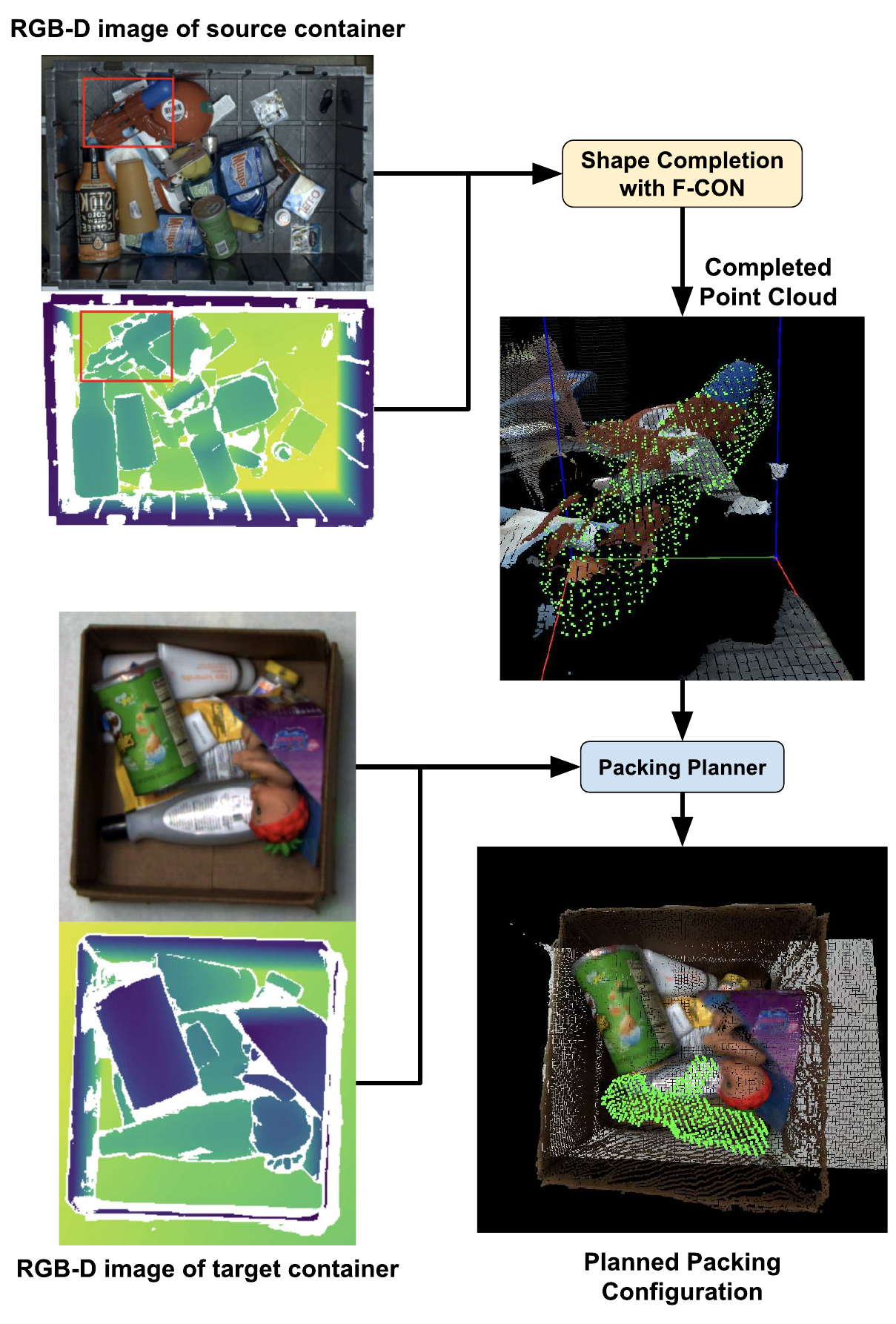}
  \vspace{-5mm}
  \caption{
  Our proposed shape completion architecture, F-CON, is trained in simulation and predicts completed point clouds for complex, unseen objects in the real world.
  Off-the-shelf packing planners can leverage F-CON for precise dense-packing in pick-and-place applications.
  }
  \vspace{-5mm}
  \label{figure_overview}
\end{figure}

Consider the requirements placed upon real-world pick-and-place systems: they need to be capable of handling a huge variety of objects, many of which will be unseen by the system prior to the moment they need to be manipulated.
In the context of dense packing, these systems need have a strong understanding of the objects' 3D geometry.
The difficulty of generalization to novel objects is exacerbated by the fact that objects are only partially observed in most applications -- for example, they are typically presented in cluttered bins where occlusions make it difficult to perceive the entire 3D shape.
Even the visible portions can pose a challenge, as many items are made of materials that cannot be easily sensed by depth cameras.

In this work, we present a shape completion model that provides the necessary perceptual understanding for dense packing systems.
Shape completion is a well-studied vision task, where the goal is typically to predict the entire 3D shape  of an object based on limited information, such as a partial point cloud.
Since this problem is inherently partially observed, shape completion models are forced to learn strong priors about 3D object geometry, making them a good fit for pick-and-place applications where the ability to deal with novel objects is essential.

Our main contributions are as follows:
\begin{enumerate}
    \item [(1)]
    We release a simulated dataset that can be used to train shape completion models for robotics applications.
    This dataset, COB-3D-v2, exhibits state-of-the-art visual realism, making it effective for sim-to-real transfer of perceptual tasks.
    It is publically available at \href{https://sites.google.com/view/fcon-packing}{our project page}.
    \item [(2)]
    We propose a 3D fully-convolutional model architecture for shape completion that performs well in the robotics domain.
    We show that our model achieves state-of-the-art performance on COB-3D-v2.
    \item [(3)]
    Through extensive real-world experiments,
    we show that our model trained on COB-3D-v2 can be combined with simple, off-the-shelf planning methods to enable state-of-the-art dense packing performance on cluttered scenes with complex, novel items.
\end{enumerate}

\section{Related Work}
\label{section_related_work}

Dense packing has been mostly studied from a planning perspective: in what order and pose should the items be placed to maximize the packed density?
Early work proposed heuristics like Deepest-Bottom-Left-First (DBLF) or Heightmap-Minimization (HM) \cite{dblf} \cite{hm}.
Besides their simplicity, these heuristics are attractive because they empirically perform well even in situations where the entire item set to be packed is not know in advance.
More recent work has attempted to learn policies using reinforcement learning (RL), exploring state/action representations, reward functions, neural network architectures, and RL algorithms that work best for this problem domain \cite{rl1} \cite{rl2} \cite{rl3} .
However, the RL-for-packing work has mostly been limited to simulated evaluations of cuboid objects, limiting its applicability to real-world systems.
Methods for 3D bounding-box estimation could help extend this work to arbitrary objects, but the imprecision of the bounding-box representation would likely result in suboptimal packing performance. Our experiments will explore this in Section IV.

Packing of complex objects in the real world has only been explored under restricted settings, such as where the ground-truth object geometry is known in advance, where the items are already singulated, or where the items only need to be packed in a 2D planar configuration \cite{real1} \cite{real2}.
These simplifications reduce the perception requirements necessary for a packing system, but do not reflect the challenges encountered in real-world applications.
In this work, we consider a more realistic setting where the items must be picked from a cluttered bin and packed into a dense 3D arrangement.
As we show in Section \ref{sec:experiments}, the strong geometric priors learned by our shape completion model enable existing planning methods to gracefully handle this difficult task.
Additionally, we explicitly evaluate the achieved packing density, which, to the best of our knowledge, has not been explored in prior work.

Methods for shape completion can be roughly categorized by the particular 3D representation that they predict.
Recent work has focused on implicit functions, which offer the best accuracy and resolution, but have limited ability for generalization  and are computationally expensive during inference   \cite{deepsdf}\cite{occ_net}.
While such methods are incredibly effective in many graphics applications, these properties make them a poor fit for robotic systems that need to handle unseen objects in low-latency applications.
Other representations like voxel grids and unstructured point clouds are more computationally tractable to work with; voxel grids tend to be more amenable to prediction with neural networks, but scale poorly to extremely high resolutions.
We will discuss how these trade-offs influence our system in Section \ref{sec:method}.

In robotics, shape completion has mostly seen attention in the context of grasping.
There have been labor-intensive attempts to collect real-world training data by taking RGB-D captures of objects with known meshes, and then use the resulting  shape completion models to plan parallel-jaw grasps on singulated objects \cite{real3}.
Later work attempted to train shape completion models on existing, generic, simulated datasets,
and used the predictions to evaluate both grasp quality and kinematic/collision feasibility during placement \cite{real2}.
However, the poor real-world performance of their shape completion model necessitated substantial focus on how the planning algorithm could reason about perceptual errors.
We evaluate that model as a baseline in Section \ref{sec:experiments}.

\section{Dense Packing with Convolutional Occupancy Models}

\subsection{Frustum-Convolutional Occupancy Networks}
\label{sec:method}

The shape completion problem is typically posed as follows: given a partial point cloud of an object, the goal is to produce a complete point cloud of the entire object surface, including invisible or occluded portions.
This is particularly amenable to robotics, where RGB-D cameras can provide partial point clouds of varying quality.
In existing benchmarks, scenes typically contain only a single object, or the partial point clouds are already segmented into objects.
However, for a real-world application where objects appear in cluttered scenes, we also require access to an instance segmentation model.
Model families such as Mask-RCNN or DETR are relatively mature and have been extensively used in robotic applications \cite{maskrcnn} \cite{detr}.

Given an RGB-D image and corresponding instance segmentation, our model predicts voxels for each object using a 3D fully-convolutional architecture.
We find that inference with voxel representations is still efficient at the resolutions we care about, and the convolution-based network architecture is the best way to impose strong inductive biases when working with structured data like RGB-D images.

\begin{figure}[thpb]
  \centering
  \includegraphics[width=0.48\textwidth]{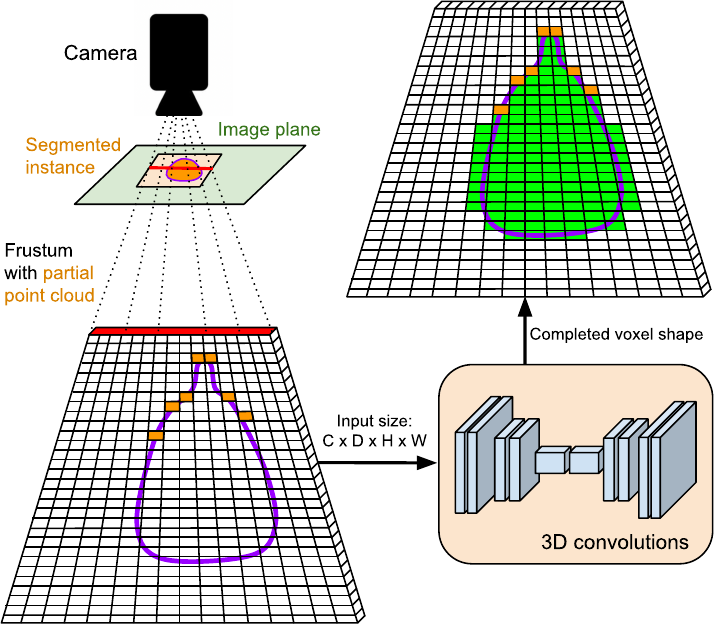}
  \caption{
  F-CON unprojects a segmented instance into a camera-aligned frustum, populates the frustum with the instance's partial point cloud (orange voxels), and then applies a series of 3D convolutions to produce a voxel grid of the completed shape (green).
  The ground-truth instance shape (purple contour) is used as supervision during training. Here we only visualize a single slice of the frustum (indicated by the red line in the image plane). In practice, F-CON unprojects the entire region-of-interest for each object.
  }
  \label{figure_frustum_arch}
  \vspace{-7mm}
\end{figure}

As illustrated in Figure \ref{figure_frustum_arch}, we first construct a trapezoidal frustum for each object.
Each frustum is the projection of its 2D bounding-box into 3D, clipped between a near plane and a far plane.
The near plane is chosen to be slightly closer to the camera than the nearest point in the partial point cloud in the object's instance mask, and the far plane
is chosen to be conservatively far based on the working volume of the scene.
We then discretize each frustum into a voxel grid, and associate a feature vector of dimension $C$ with each voxel, resulting in a feature volume of shape $C \times D \times H \times W$ for each instance.
For all experiments, we chose $D = 96, H = W = 64$ for a favorable trade-off between inference speed and performance.
The voxels are trapezoidal, but they are aligned with the camera viewpoint (for a given $h, w$ coordinate, every voxel along the $D$ dimension is on the same ray entering the camera, and projects to the same pixel in the image plane), and they are spaced linearly along the $D$ dimension between the near and far plane.
For each point in the partial point cloud, we fill the corresponding voxel with the RGB color and a binary indicator for the instance mask ($C = 4$).
The camera-centric and object-centric properties of this scheme encourage robustness to different camera viewpoints, scene composition, and object sizes, which improves the sim-to-real transfer performance.
A similar frustum-based scheme was used by Mesh-RCNN; however, they predict voxels jointly with instance masks, and do not condition on a partial point cloud \cite{meshrcnn}.
The latter is desirable in applications where depth is not available during inference, but depth cameras are already ubiquitous in robotics and can substantially improve performance.

The initial $C \times D \times H \times W$ feature volume is passed through a 3D-convolutional UNet to produce an updated feature volume of the same shape.
We then use a 3D-conv layer to reduce the $C$ dimension to 1, and then refine the resulting $D \times H \times W$ feature map with a 2D UNet (where the $D$ dimension is treated as the channel dimension).
This is an efficient way to increase the expressiveness of the model, since 2D convolutions are cheaper than their 3D counterparts.
Each element of the 2D UNet's output (still $D \times H \times W$) is the scalar probability that the corresponding voxel is occupied by the object's completed shape.
During training, we label each voxel as positive if it is contained inside the object's ground-truth mesh. 
Each voxel is supervised independently using a class-balanced binary cross-entropy loss.
During inference, we extract meshes from the voxel predictions using Marching Cubes \cite{marching_cubes} and 
sample points uniformly from the surface to obtain an unstructured point cloud.
For more details, see \href{https://github.com/nikhilmishra000/fcon/}{our implementation}.

We call this model F-CON (\textit{F}rustum-\textit{C}onvolutional \textit{O}ccupancy \textit{N}etwork). 
In the following sections, we discuss the dataset used to train it, and how we utilize it for dense packing in the real world.

\subsection{Simulated Dataset}

\begin{figure}[pbh!]
  \centering
  \includegraphics[width=0.41\textwidth]{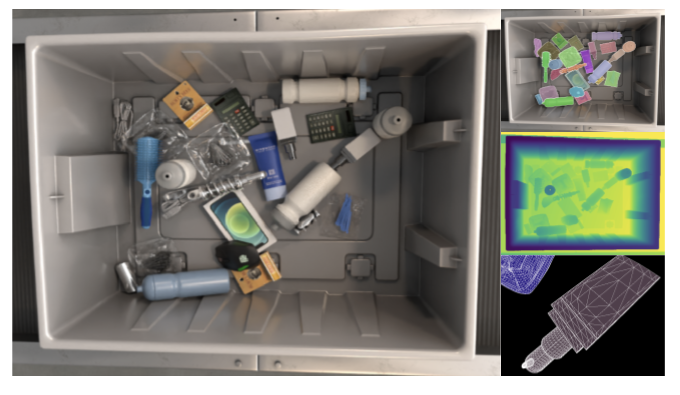}
  \vspace{-4mm}
  \caption{
  COB-3D contains high quality RGB renderings (left), instance masks (top right), and depth maps (middle right).
  In this work, we released a new version, COB-3D-v2, that also includes meshes for each instance (bottom right).
    It can be downloaded from \href{https://sites.google.com/view/fcon-packing}{our project page}.
  }
  \label{figure_cob3d}
\end{figure}

COB-3D is a simulated dataset of common objects in bins, arranged in realistic yet challenging configurations \cite{bbox_paper}.
As illustrated in Figure \ref{figure_cob3d}, the dataset contains roughly 7000 scenes of high-quality RGB renderings along with ground-truth camera calibrations, instance masks and point clouds.
Each scene contains up to 30 objects, which are thrown into a bin using physics simulation.
For robust sim-to-real transfer, the object sizes, camera parameters, and scene lighting are all randomized.
In this work, we release a new version of this dataset, COB-3D-v2, with ground-truth meshes and poses of each object instance. 
This addition allows shape completion models such as F-CON to be trained on COB-3D-v2.
For more details about the dataset format, see Appendix \ref{sec:dataset_format}.
Example scenes are visualized in Appendix \ref{sec:dataset_examples}.

Note that neither COB-3D nor COB-3D-v2 have object categories (all objects belong to a single category). 
This differs from prior work in shape completion, where common practice is to either train category-specific models or condition on the object category.
However, we find that the lack of categories is more representative of real-world settings with novel objects, which may belong to arbitrary novel categories, or may be hard to categorize in the first place.

We trained F-CON for 125 epochs on COB-3D-v2, which took about 4 GPU-days.
 We used a batch size of 32 scenes and the Adam optimizer with default hyperparameters ($\alpha = 10^{-3}, \beta_1 = 0.9, \beta_2 = 0.999, \epsilon = 10^{-8}$).
During training, we also randomize the near and far planes for each instance, which improves robustness after sim-to-real transfer.
In Section \ref{sec_sim_eval}, we evaluate F-CON against baselines for shape completion on COB-3D-v2.

\subsection{Dense Packing with F-CON}
\label{sec_planner}

Given a trained shape completion model, there can be different ways to utilize it in a dense-packing system.
To highlight F-CON's perceptual capabilities, we opted for a simple planning pipeline that allows us to use off-the-shelf methods from prior work.
As described in Algorithm \ref{alg_planner}, we determine a grasp pose $g^* \in SE(3)$ and placement pose $q^* \in SE(3)$  from the following inputs:
\begin{itemize}
    \item 
    A height-map $H[\cdot, \cdot]$ for the target container, which is computed from the captured depth map.
    The container is discretized into rectangular cells, where $H[u, v] = z$ if the highest sensed point in cell $(u, v)$ is distance $z$ from the container bottom.
    \item 
    A set of candidate grasp poses $G = \{ g^{(1)}, \dots, g^{(N} \}_{i=1}^{N}, g^{(i)} \in SE(3)$.
    These can be generated using any method and for any grasping modality (suction, parallel-jaw, etc).
    \item 
    Completed point clouds $O = \{ o^{(i)} \}_{i=1}^{N}$ for each grasped object.
    Each $o^{(i)} \in \mathbb{R}^{K \times 3}$ is expressed in the $g^{(i)}$ frame.
    \item 
    A cost function $C(g, q) \rightarrow \mathbb{R}$ that evaluates the packing quality of a candidate grasp $g$ and placement $q$.
\end{itemize}

Given the returned grasp and placement poses, we use a scripted trajectory planner such that the robot's gripper retracts linearly upwards from the grasp pose, and descends linearly downwards during placement.

\begin{figure}[tph!]
  \centering
  \includegraphics[width=0.5\textwidth]{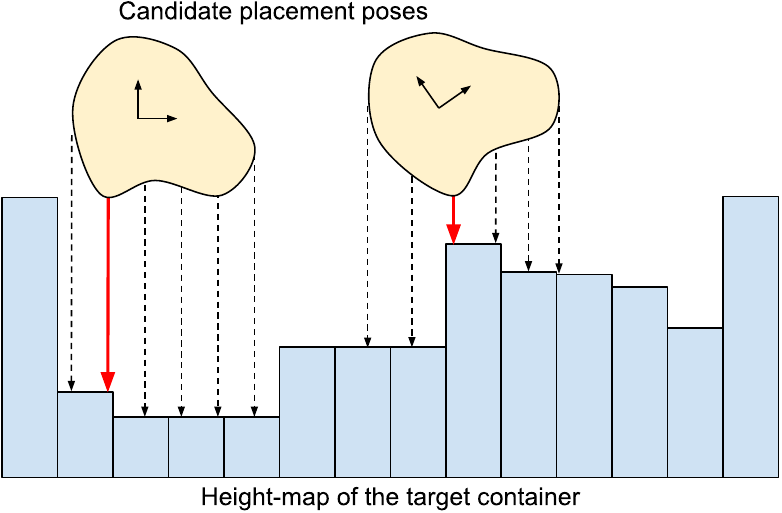}
  \caption{
  A 2D-slice of the lowest-placeable-height computation from Algorithm \ref{alg_planner}.
  For a candidate placement pose, every point in the completed shape (yellow) is projected (black/red arrows) onto the target container's height-map (blue).
  The shortest projection distance (red arrow) determines the height at which the object can be placed for that cell and orientation.
  In this example, the left candidate results in a lower placement than the right.
  }
  \label{figure_placeable_height}
\end{figure}

\begin{algorithm}[h]
\caption{A simple planner for dense packing}
\label{alg_planner}
\SetKwInOut{Input}{Input}
\Input{Height-map $H$ for the target container \\
Candidate grasps $G = \{ g^{(1)}, \dots, g^{(N)} \}$ \\
Object point clouds $O = \{ o^{(1)}, \dots, o^{(N)} \}$ \\
Placement cost function $C(g, q) \rightarrow \mathbb{R}$
}

Initialize $g^* \leftarrow \text{null}, q^* \leftarrow \text{null}, c^* \leftarrow \infty$\;
\For{k = 1, \dots, M}{
Sample a grasp $g^{(k)}$ from $G$\;
Sample a placement cell $(u^{(k)}, v^{(k)})$ inside $H$\;
Sample a placement orientation $R^{(k)}$\;
Compute the lowest placeable height $z^{(k)}$ (see Figure \ref{figure_placeable_height}) for object $o^{(k)}$, when centered at $H[u^{(k)}, v^{(k)}]$ and rotated by $R^{(k)}$\;
Compute the gripper pose $q^{(k)}$ corresponding to $(i^{(k)}, j^{(k)}, z^{(k)}, R^{(k)})$\;
Evaluate $c^{(k)} \leftarrow C(g^{(k)}, q^{(k)})$\;
\If{$c^{(k)} < c$ \textbf{and} $\text{MotionFeasible}(g^{(k)}, q^{(k)}) $}{
    $g^* \leftarrow g^{(k)}, q^* \leftarrow q^{(k)}, c^* \leftarrow c^{(k)}$\;
}
}
\KwResult{
Best placement $q^*$, corresponding grasp $g^*$
}
\end{algorithm}
\vspace{2mm}

This framework neatly encapsulates existing planning methods like DBLF and HM.
Using $C(g, q) = q_z + \epsilon\cdot(q_x + q_y)$, for $0 < \epsilon << 1$, yields the DBLF planner.
The HM planner estimates the height-map $H'(q)$ that would result from placement $q$, and then uses $C(g, q) = \sum_{u, v} H'(q)[u, v] - H[u, v]$.
For more details about how $H'(q)$ is computed, see \cite{hm}.
Combining F-CON with a model-based RL method (such as by extending the simulated packing work discussed in Section \ref{section_related_work}) could potentially result in a better cost function than either DBLF or HM, as well as a better sampler than the uniform one that we use.
However, to focus on evaluating F-CON, we defer such explorations to future work.

Unlike some prior work, we do not consider re-grasping, where already-packed items may be removed in order to achieve a better arrangement.
Re-grasping allows the system to mitigate the effects of perceptual mistakes made in the preceding timesteps, which confounds our evaluation of shape completion models.

\section{Experiments}
\label{sec:experiments}

We conducted extensive real-world experiments to answer the following questions:
\begin{itemize}
    \item [(1)]
    How effective is F-CON at shape completion, as trained and evaluated on the realistic-but-simulated scenes in COB-3D-v2? 
    \item [(2)]
    To what extent does F-CON address the perceptual difficulties surrounding dense packing, as evaluated on novel objects in cluttered real-world scenes?
\end{itemize}

\begin{table*}[th!]
\caption{Results on COB-3D-v2}
\vspace{-2mm}
\label{table_cob3d_chamfer}
\begin{center}
    \begin{tabular}{c|c c c c c |c c c}
        &  \multicolumn{5}{|c|}{Shape completion} & \multicolumn{3}{c}{3D bounding-box estimation} \\
        Method & CD-L1 ($\downarrow$) & CD-L2 ($\downarrow$) & F1$^{0.1} (\uparrow)$ & F1$^{0.3} (\uparrow)$ & F1$^{0.5} (\uparrow)$ & Box-IoU ($\uparrow$) & Box-IoG ($\uparrow$) & Box-F1 ($\uparrow$)\\
        \hline
        AR-bbox \cite{bbox_paper} & - & - & - & - & - & 0.6296 & 0.7877 & 0.6999 \\
        PCN \cite{pcn} & 0.9835 & 0.5800 & 0.0729 & 0.5460 & 0.7968 & 0.5374 & 0.8002 & 0.6460 \\
        PoinTr \cite{pointr} & 0.4857 & 0.1582 & 0.3721 & 0.8717 & 0.9555 & 0.5874 & \textbf{0.8394} & 0.6984 \\
        \hline
        F-CON (ours) & \textbf{0.4229} & \textbf{0.1157} & \textbf{0.4664} &  \textbf{0.8928} & \textbf{0.9600} & \textbf{0.6809} & 0.7686 & \textbf{0.7485}\\
        \hline
    \end{tabular}
\end{center}
\vspace{-7mm}
\end{table*}

\subsection{COB-3D-v2 Evaluation}
\label{sec_sim_eval}

\begin{figure}[bph!]
    \centering
    \includegraphics[width=0.15\textwidth]{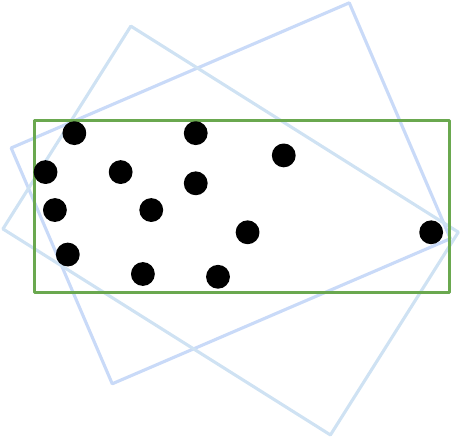}
    \caption{
    Fitting a bounding box around a point cloud (black dots).
    We sample several candidate boxes that contain all the points (blue), and then take the one with minimum volume (green).
    Notice that the single point on the far right substantially influences the dimensions of the fitted box.
    }
    \label{fig_bbox_sampling}
    \vspace{-4mm}
\end{figure}

To benchmark shape completion on COB-3D-v2, we considered several metrics, following  prior work:
\newpage
\begin{itemize}
\item 
\textbf{Chamfer distance}: This is the standard metric for comparing unstructured point clouds in benchmarks such as ShapeNet \cite{shapenet}.
Given two point clouds $X$ and $Y$, the Chamfer distance (CD) is computed as follows:

\small
\begin{align*}
&\text{CD}(X, Y) = \frac{1}{|X|} \sum_{x \in X} \min_{y \in Y} \| x - y \|_2^2
+ \frac{1}{|Y|} \sum_{y \in Y} \min_{x \in X} \| y - x \|_2^2
\end{align*}
\normalsize
Often, an L1-variant (CD-L1), where the $\|\cdot\|_2^2$ norms are replaced by $\| \cdot \|_1$, is reported alongside the traditional Chamfer-L2 distance  (CD-L2).
\vspace{2mm} 

\item 
\textbf{$\text{F1}^\tau$}: 
For a given distance threshold $\tau$, predicted point
cloud $X$ and ground-truth point cloud $Y$ , $\text{F1}^\tau(X, Y)$
is the harmonic mean of the precision at $\tau$ (fraction
of points in $X$ that are within $\tau$ of some point in $Y$ )
and the recall at $\tau$ (fraction of points in $Y$ that are
within $\tau$ of some point in $X$). This metric is usually
considered alongside the Chamfer distance in shape
completion benchmarks because it is less sensitive to
outliers, and is typically reported at varying values of $\tau$.
\vspace{2mm} 

\item 
\textbf{Box IoU, IoG, F1}:
Given a completed point cloud, a 3D bounding-box can be fit around it and compared it to the ground-truth bounding-box, using metrics from 3D bounding-box estimation.
In contrast with Chamfer distances, and especially $\text{F1}^\tau$, we find that bounding-box metrics are very sensitive to outliers.
However, they may be more representative of packing performance, since outliers can cause a packing system to believe that an item cannot fit in a pose where it actually could have.
To fit a bounding box around a point cloud, we sample rotations uniformly at random in quaternion space \cite{bbox_sampling}, compute the axis-aligned dimensions of the enclosing box in each sampled rotation frame, and finally choose the sampled box with the smallest volume (see Figure \ref{fig_bbox_sampling} for a visualization).
Using this scheme, we report IoU, IoG, and F1: IoU is the standard metric for bounding-box estimation, IoG (intersection-over-ground-truth) is a form of recall to complement IoU, and F1 trades off between the two \cite{bbox_paper}.
Note that IoG is not particularly meaningful in isolation, since perfect IoG can be achieved by simply predicting arbitrarily large bounding-boxes.

\end{itemize}

We considered the following methods as baselines to F-CON.
Like prior work in shape completion for grasping, we did not consider implicit functions: they must be queried extremely densely to extract surface geometry, and often require test-time optimization, making them too computationally expensive during inference for use in a real-world system \cite{deepsdf} \cite{occ_net}.

\begin{itemize}
    \item 
    PCN \cite{pcn} has been used in prior work for grasp and placement planning \cite{real2}.
    It uses an encoder that embeds a partial cloud into a latent space, and a decoder that constructs the completed point cloud from the latent vector.
    Both encoder and decoder use PointNets \cite{pointnet} to operate directly on unstructured point clouds directly, and they are trained end-to-end using the Chamfer-L2 distance as the loss function.
    To train PCN on COB-3D-v2, we normalize each instance's partial point cloud using its frustum, decode the completed point cloud in the normalized coordinates, and then transform it back to the original space.
    For a more fair comparison with F-CON, we also improve upon the original architecture by concatenating RGB and instance masks with the partial point cloud.

    \item 
    PoinTr \cite{pointr} uses a similar encoder-decoder framework to PCN, but substantially improves the architecture, primarily by using Transformers \cite{transformer}.
    It achieves state-of-the-art performance on many shape completion benchmarks even outside of robotics.
    We train PoinTr using the same normalization scheme, additional inputs, and loss function as PCN.

    \item 
    The autoregressive bounding-box model (AR-bbox) that accompanied the original COB-3D release has been shown to perform well on 3D bounding-box estimation \cite{bbox_paper}.
    Since bounding boxes have been used as a simplified state representation in prior packing work (as discussed in Section \ref{section_related_work}), we also consider this model as a baseline, but only evaluate it on bounding-box metrics.
\end{itemize}

Both Chamfer distance and F1$^\tau$ generally depend on both the scaling and density of the point clouds.
Following prior work \cite{meshrcnn}, we scale all point clouds such that the longest edge of the ground-truth bounding-box has length 10.
The ground-truth point clouds are generated by sampling 16384 points uniformly from the mesh surface.
For F-CON, we sample points in the same way, but from the meshes obtained via Marching-Cubes.
PCN and PoinTr always output a fixed number of points, so we train them accordingly to predict 16384 points.

\begin{figure}[tph!]
  \centering
  \includegraphics[width=0.475\textwidth]{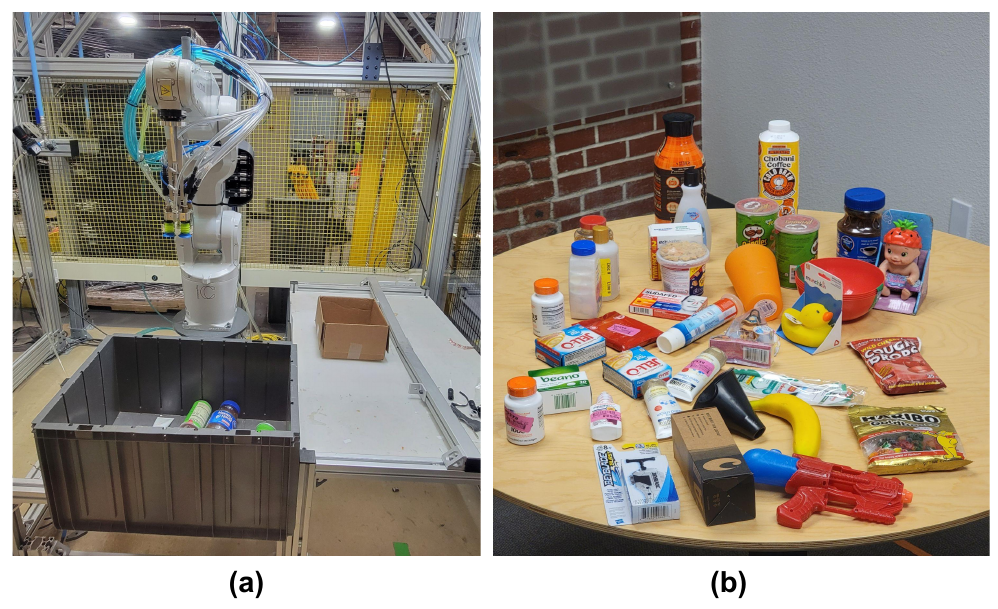}
  \vspace{-2mm}
  \caption{
  (a) Our robot picks items from the cluttered bin in the front, and packs them into the cardboard box on the the table.
  (b) The complete item set used in our experiments.
  }
  \label{figure_physical_setup}
\end{figure}

In Table \ref{table_cob3d_chamfer}, we see that F-CON outperforms the baseline methods across all shape completion metrics.
We find these results particularly compelling given that F-CON is not trained to minimize Chamfer distance, unlike PCN and PoinTr.
We also note that F-CON performs well on 3D bounding-box metrics, even though PCN and PoinTr do not.
Qualitatively, we observe PCN and PoinTr are prone to outliers in their predicted point clouds -- this is well-reflected in their F1$^\tau$ and IoG scores.

\subsection{Real-World Dense Packing}

We designed real-world experiments to mimic typical order fulfillment applications, where a robot must pack items from a cluttered bin (or often, multiple bins) into a smaller container, like a cardboard box. 
As shown in Figure \ref{figure_physical_setup}, we used an ABB-1200 with a 5-cup suction gripper, with RGB-D cameras mounted above both containers.
The item set consists of a variety of household objects of diverse shapes and categories.
In total, we have 35 objects, which are unseen by all models (since all models are trained purely in simulation on COB-3D-v2).

Recall that the goal of dense packing is to minimize the volume that the packed objects occupy.
Thus, we evaluate our system by directly estimating this quantity at the end of each episode, using a scheme inspired by the HM heuristic. 
To the best of our knowledge, no prior work has evaluated real-world packing quality with a continuous volumetric measure (a common practice is simply to check whether all items were successfully placed in the container).
Using the target container's height-map, as defined in Section \ref{sec_planner}, we can estimate the total volume occupied by the packed objects via numerical integration over the cells of the height-map.
The HM planner minimizes the change in this quantity for each item to be packed; while tractable and easy to implement, this is generally not optimal when considering the entire episode.

For each shape completion model from Section \ref{sec_sim_eval}, we use the planner from Algorithm \ref{alg_planner} using either DBLF or HM as the cost function $C(g, q)$, height-map cells of 1 mm $\times$ 1 mm, and a sample size of $M = 4096$ placements.
Since F-CON operates in an object-centric manner, its inference time scales with the number of objects in the scene.
For a large scene (16 objects), it takes about 25 milliseconds on an NVIDIA Quadro RTX 600.
The entire planning process (perception, sampling placements, scoring, motion planning) takes about 300 milliseconds.

Within each trial, we select a subset ranging from 5 to 15 items uniformly at random from the overall item set, and arrange them chaotically in the source container.
Each trial consists of one episode for each (\textit{shape completion}, \textit{planner}) configuration, wherein the system packs the sampled items one-by-one into the target container.
The episode ends when either all items are placed, one is placed that causes the container to overflow, or the system cannot find an overflow-free plan.
At the end of each episode, we estimate the packed volume using the height-map integration scheme discussed in the previous paragraph.
Finally, we measure human performance by quickly packing the items by hand (taking 20 seconds or less), and measuring the packed volume in the same manner.

In Table \ref{table_packing}, we show the performance of each shape completion model paired with different off-the-shelf packing planners, alongside human performance, across 50 trials.
We report the mean and standard error for the following metrics:
\begin{itemize}
    \item 
    \textbf{Success Rate}: the fraction of episodes where all items were successfully packed at the end of the episode.
    
    \item 
    \textbf{Packed Volume}: the volume measured by height-map integration at the end of the episode.
    We express this value as a fraction of the total container volume.
    In episodes that are not successes, we record a value of 1.0, which is the worst possible score and corresponds to using the entire container.
\end{itemize}

\begin{figure*}[tph!]
  \centering
  \includegraphics[width=1.0\textwidth]{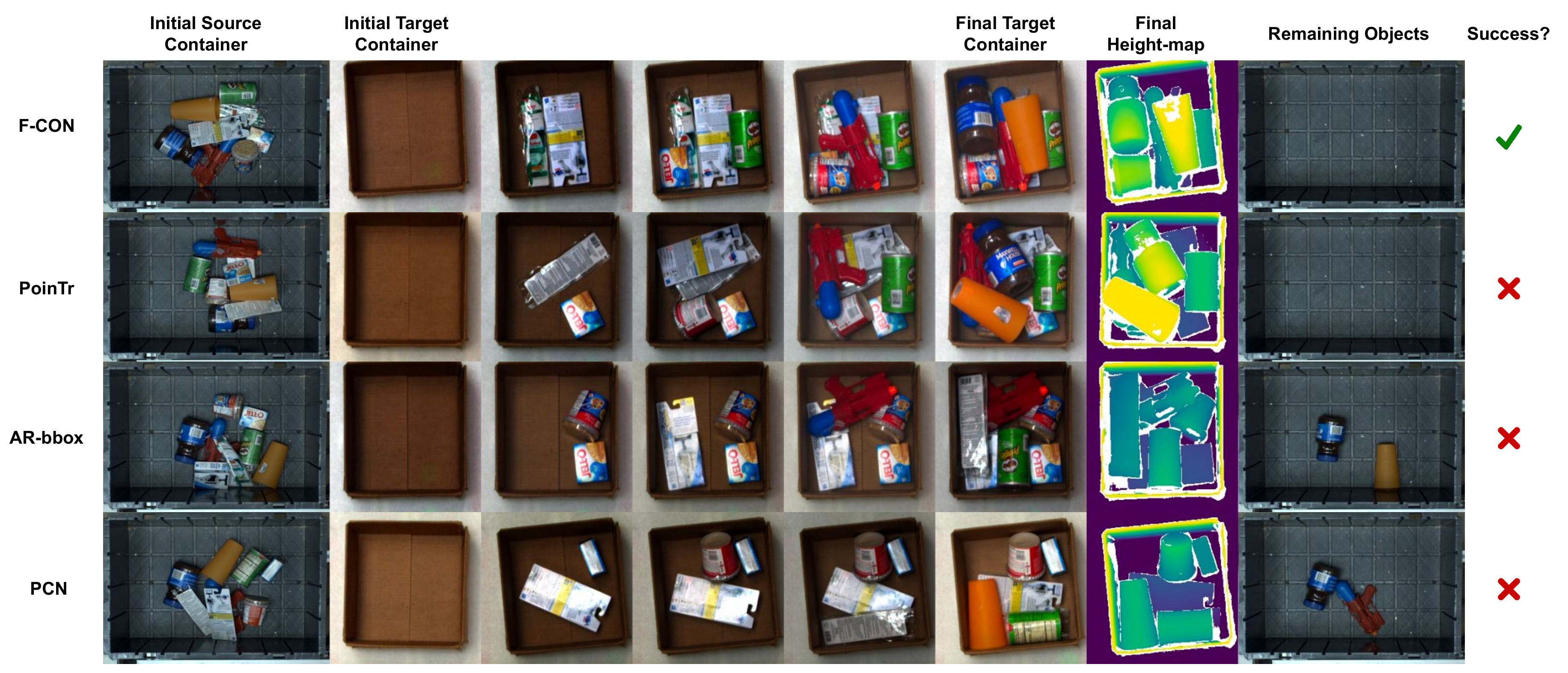}
  \caption{
  Representative episodes from our real-world packing experiments.
  In each row, we conduct one episode using the listed shape completion model and HM as the planning method.
  All rows use the same item set.
  The progress within the episode can be seen from left to right, along with the final height-map.
  With F-CON (row 1), all of the items can be packed densely.
  Other methods cause the packing container to overflow (row 2) or they cannot pack all the items (rows 3, 4).
  }
  \label{figure_packing_episodes}
\end{figure*}

For all planners, F-CON substantially outperforms the other shape completion methods, demonstrating its efficacy for real-world dense packing.
In Figure \ref{figure_packing_episodes}, we visualize representative episodes for a qualitative understanding.
With other methods, the system often cannot find a suitable placement pose or causes the target container to overflow.
The former typically results from overestimation of the object's size, which is consistent with the F1$^\tau$ and IoG results discussed in Section \ref{sec_sim_eval}.

\begin{table}[bph!]
\caption{Real-World Dense Packing}
\vspace{-3mm}
\label{table_packing}
\begin{center}
    \begin{tabular}{c c|c c}
Shape Completion                 & Planner  & Packed Volume ($\downarrow$)     & Success Rate ($\uparrow$)       \\
\hline
AR-bbox \cite{bbox_paper}        & DBLF     & $0.698\pm0.047$                    & $0.48\pm0.071$                    \\
PCN \cite{pcn}                   & DBLF     & $0.909\pm0.035$                    & $0.12\pm0.046$                    \\
PoinTr \cite{pointr}             & DBLF     & $0.570\pm0.046$                    & $0.66\pm0.067$                    \\
F-CON (ours)                            & DBLF     & $\textbf{0.461}\pm0.041$                    & $\textbf{0.80}\pm0.057$                    \\
\hline
AR-bbox \cite{bbox_paper}        & HM       & $0.762\pm0.045$                    & $0.38\pm0.069$                    \\
PCN \cite{pcn}                   & HM       & $0.664\pm0.049$                    & $0.50\pm0.071$                    \\
PoinTr \cite{pointr}             & HM       & $0.535\pm0.039$                    & $0.76\pm0.060$                    \\
F-CON (ours)                            & HM       & $\textbf{0.440}\pm0.032$                    & $\textbf{0.88}\pm0.046$                    \\
\hline
Human                            & Human    & $0.357\pm0.012$                    & $1.00\pm0.000$                    \\
    \hline
    \end{tabular}
\end{center}
\end{table}

\newpage
\section{Conclusion}

We presented F-CON, a voxel-based shape completion model with strong inductive biases, and validated it on the highly-realistic simulated dataset COB-3D-v2.
We then conducted extensive experiments in the real-world, and showed that the strong geometric priors learned by F-CON can enable dense-packing of complex, unseen items in chaotic, cluttered scenes, without any real-world training.
Although using F-CON results in substantially better performance that baseline shape completion methods, one shortcoming of our packing system is the simplicity of the planning methods we used.
Combining F-CON with RL methods to obtain learned packing policies is an exciting direction for future work, and we hypothesize that this could close the gap with respect to human performance.






\newpage 
\onecolumn
\appendix

\subsection{COB-3D-v2 Dataset Format}
\label{sec:dataset_format}

COB-3D-v2 contains 6955 scenes in total (6259 train, 696 val).
For each scene, we provide the following:

\small
\begin{verbatim}
  |-- rgb:                    The rendered RGB image. Shape (3, H, W), dtype float32.
                              Values scaled to [0, 1].

  |-- intrinsic:              The camera intrinsics. Shape (3, 3), dtype float32.

  |-- depth_map:              The rendered depth map corresponding to `rgb`. 
                              Shape (H, W), dtype float32.

  |-- normal_map:             The rendered normal map corresponding to rgb`.  
                              Shape (3, H, W), dtype float32.

  |-- near_plane:             The minimum depth value of the scene's working volume.
                              Scalar, float32.
  
  |-- far_plane:              The maximum depth value of the scene's working volume.
                              Scalar, float32.

  |-- segm/
          |-- boxes:          2D bounding boxes for each object in the scene.
                              Shape (N_objects, 4), dtype float32. 
                              These are pixel coordinates relative to `rgb`. 
                              The box format is `[x_low, y_low, x_high, y_high]`

          |-- masks:          Binary masks for each object in the scene. 
                              Shape (N_objects, H, W), dtype bool.

          |-- amodal_masks:   Amodal instance masks for each object.
                              Shape (N_objects, H, W), dtype bool.

  |-- bbox3d/
            |-- poses:        The pose of each object's 3D bounding box, as a 4x4 matrix.
                              This is the transform from the bbox frame to the camera frame.
                              Shape (N_objects, 4, 4), dtype float32.

            |-- dimensions:   The dimensions of each 3D bounding box.
                              Shape (N_objects, 3), dtype float32.

            |-- corners:      The corner points of each 3D bounding box, in the camera frame.
                              Shape (N_objects, 8, 3), dtype float32.


  |-- mesh_ids:               The mesh_id of each object. List[str], length N_objects.

  |-- obj_poses/
               |-- poses:     The pose of each mesh, as a 4x4 matrix.
                              This is the transform from the mesh frame to the camera frame.
                              Note that the mesh frame does not necessarily equal the bbox frame!
                              Shape (N_objects, 4, 4), dtype float32.

               |-- scales:    The scale of each mesh. Shape (N_objects, 3), dtype float32.

  |-- voxel_grid/
                |-- voxels:   The surface of each mesh, extracted into a voxel grid.
                              Shape (N_objects, n_voxels, n_voxel, n_voxels), dtype bool.

                |-- extents:   The extents of each object's voxel grid.
                              `voxels[i]` span the cuboid `[-extents[i], extents[i]]`,
                              in the object frame `obj_poses/poses[i]`.
                              Shape (N_objects, 3), dtype float32.
\end{verbatim}
\normalsize

\newpage
\subsection{COB-3D-v2 Examples}
\label{sec:dataset_examples}

The following pages exhibit some representative scenes from COB-3D-v2, showcasing the visual quality and diversity of the dataset.
In each row, the left column is the rendered RGB, and the right column is the rendered depth map.

\begin{figure*}[tph!]
  \centering
  \includegraphics[width=1.0\textwidth]{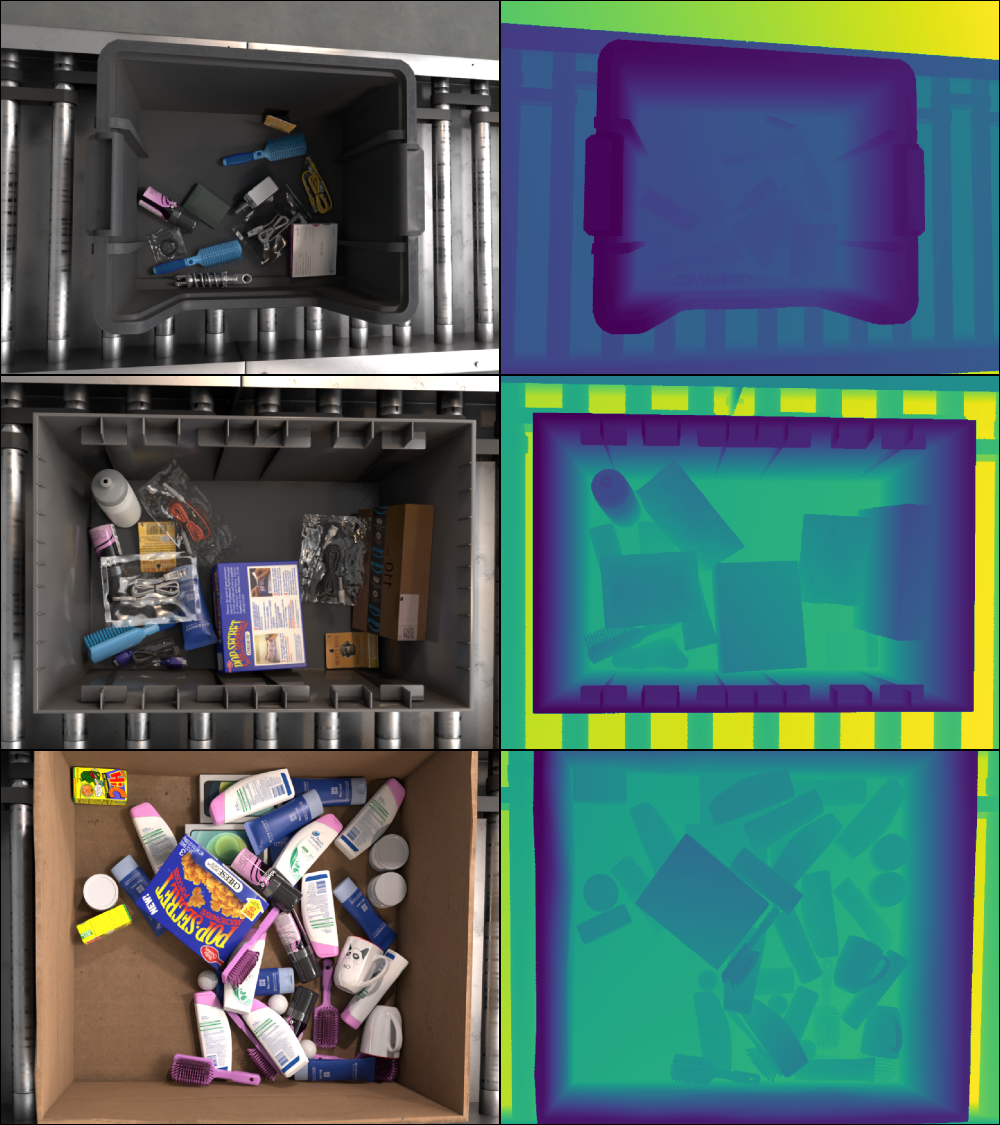}
\end{figure*}

\newpage
\begin{figure*}[tph!]
  \centering
  \includegraphics[width=1.0\textwidth]{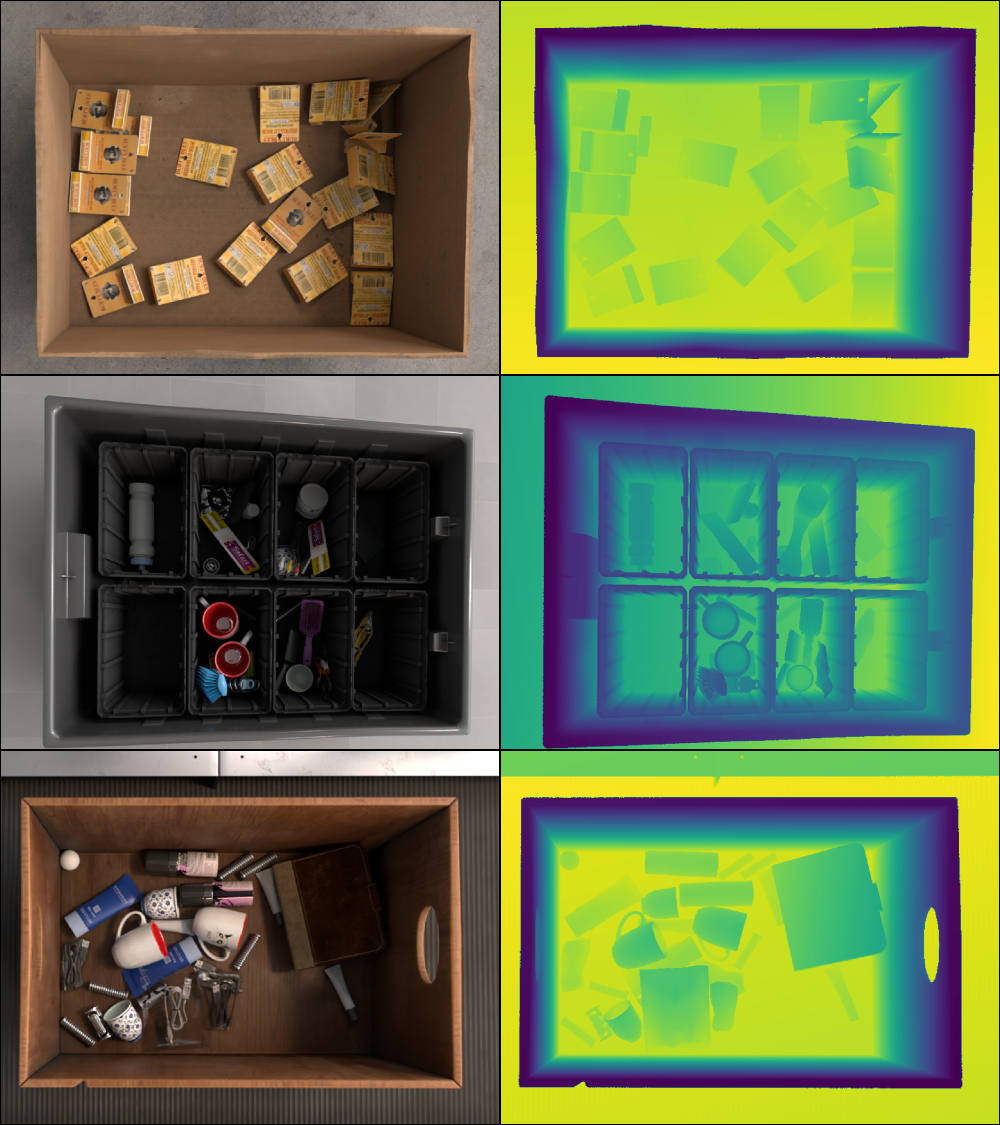}
\end{figure*}

\newpage
\begin{figure*}[tph!]
  \centering
  \includegraphics[width=1.0\textwidth]{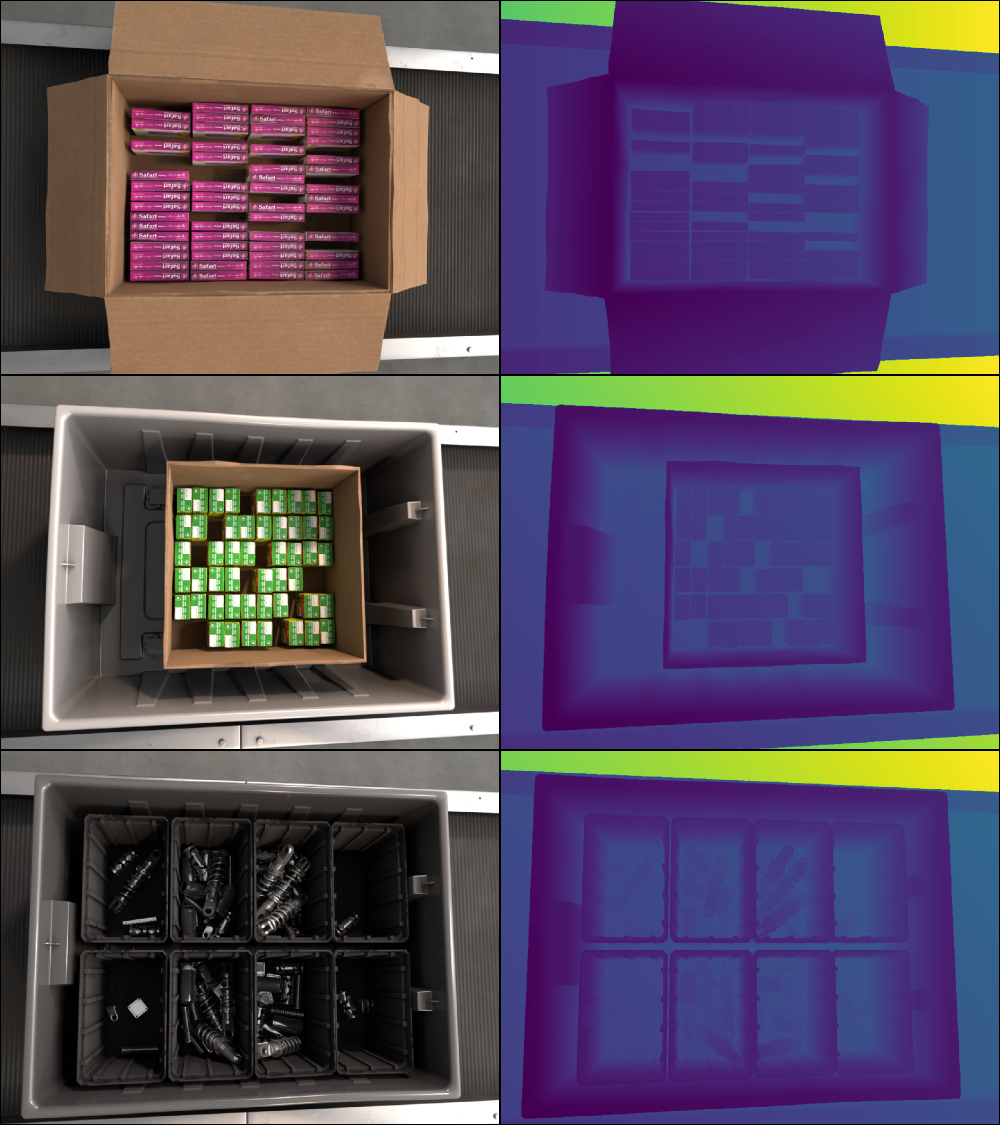}
\end{figure*}

\newpage
\begin{figure*}[tph!]
  \centering
  \includegraphics[width=1.0\textwidth]{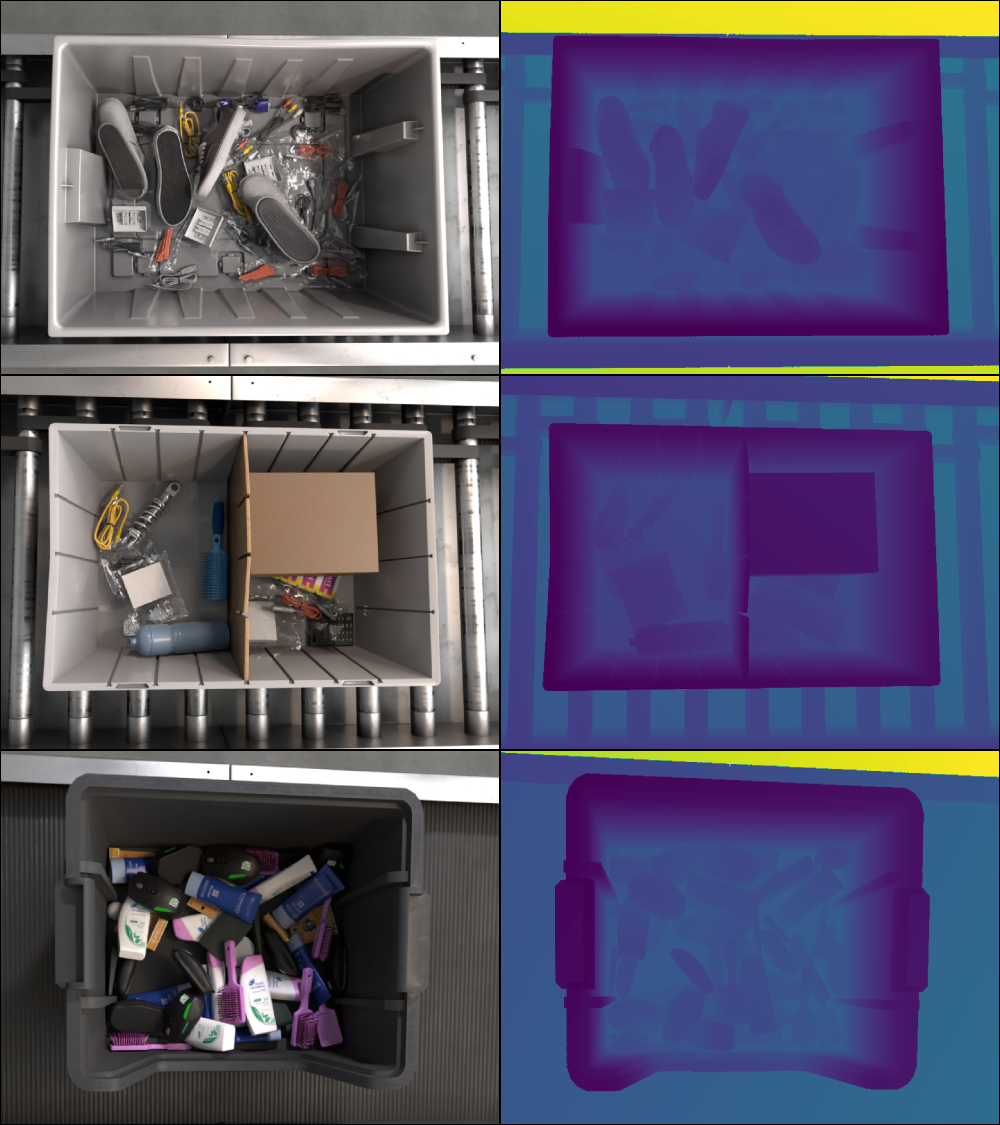}
\end{figure*}

\newpage
\begin{figure*}[tph!]
  \centering
  \includegraphics[width=1.0\textwidth]{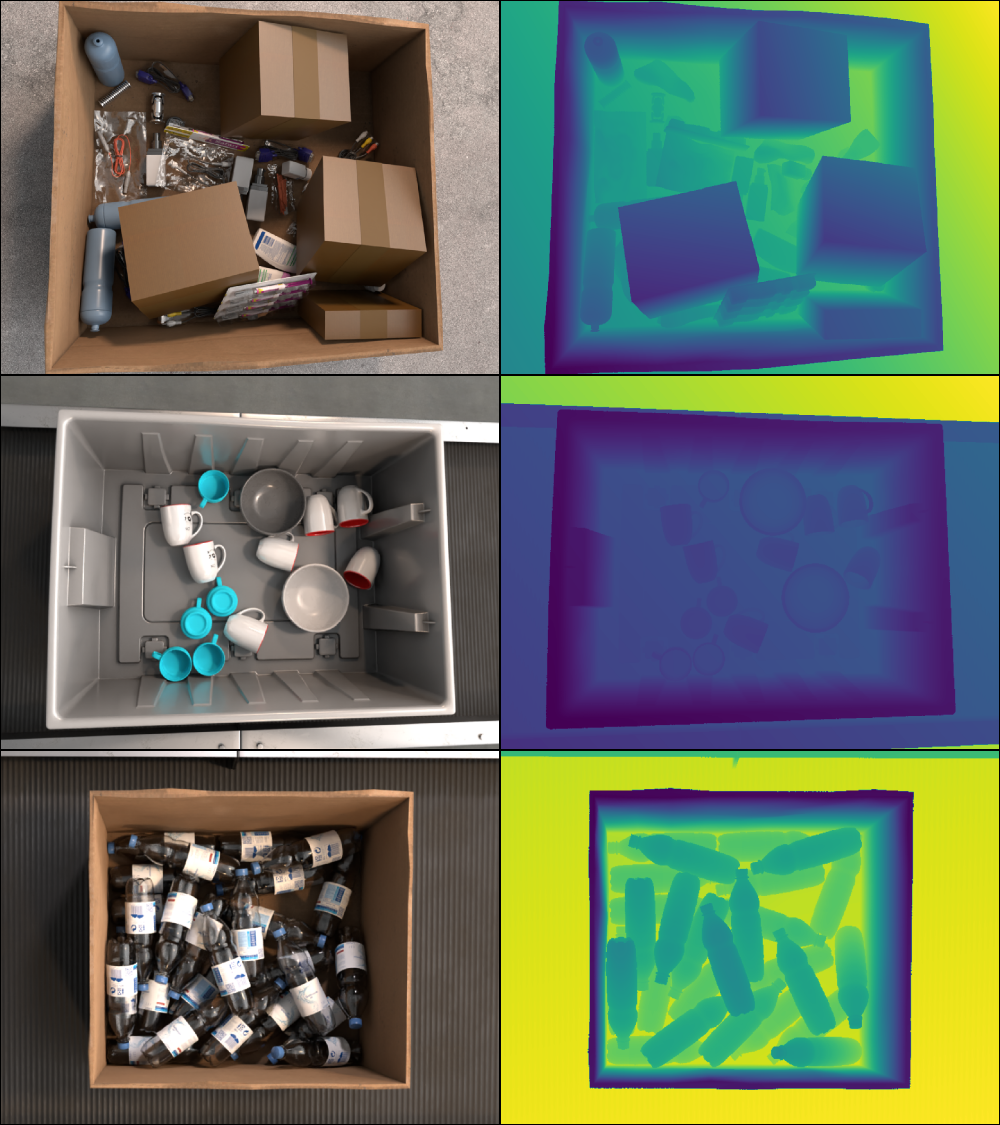}
\end{figure*}


\begin{thebibliography}{99}

\bibitem{dblf}
Wang, Lei, et al. "Two natural heuristics for 3D packing with practical loading constraints." Trends in Artificial Intelligence: 11th Pacific Rim International Conference on Artificial Intelligence, 2010.

\bibitem{hm}
Wang, Fan, and Kris Hauser. "Stable bin packing of non-convex 3D objects with a robot manipulator." International Conference on Robotics and Automation (ICRA), 2019.

\bibitem{rl1}
Kundu, Olyvia, Samrat Dutta, and Swagat Kumar. "Deep-pack: A vision-based 2d online bin packing algorithm with deep reinforcement learning." International Conference on Robot and Human Interactive Communication (RO-MAN), 2019.

\bibitem{rl2}
Yang, Shuo, et al. "Heuristics Integrated Deep Reinforcement Learning for Online 3D Bin Packing." IEEE Transactions on Automation Science and Engineering, 2023.

\bibitem{rl3}
Jia, Jie, Huiliang Shang, and Xiong Chen. "Robot Online 3D Bin Packing Strategy Based on Deep Reinforcement Learning and 3D Vision". International Conference on Networking, Sensing and Control (ICNSC), 2022.

\bibitem{real1}
Wang, Fan, and Kris Hauser. "Dense robotic packing of irregular and novel 3D objects". IEEE Transactions on Robotics 38.2, 2021.

\bibitem{real2}
Gualtieri, Marcus, and Robert Platt. 
"Robotic pick-and-place with uncertain object instance segmentation and shape completion". IEEE Robotics and Automation Letters 6.2, 2021.

\bibitem{real3}
Varley, Jacob, et al. "Shape completion enabled robotic grasping". International Conference on Intelligent Robots and Systems (IROS), 2017.


\bibitem{bbox_paper}
YuXuan Liu, et al.
"Autoregressive Uncertainty Modeling for 3D Bounding Box Prediction".
European Conference on Computer Vision (ECCV), 2022

\bibitem{meshrcnn}
Gkioxari, Georgia, Jitendra Malik, and Justin Johnson. "Mesh R-CNN". International Conference on Computer Vision (ICCV), 2019.

\bibitem{pcn}
Yuan, Wentao, et al. "PCN: Point completion network." International Conference on 3D Vision (3DV), 2018.

\bibitem{maskrcnn}
He, Kaiming, et al. "Mask R-CNN". International Conference on Computer Vision (ICCV), 2017.

\bibitem{detr}
Carion, Nicolas, et al. "End-to-end object detection with transformers". European Conference on Computer Visions (ECCV), 2020.

\bibitem{shapenet}
Chang, Angel X., et al. "Shapenet: An information-rich 3d model repository". arXiv preprint arXiv:1512.03012 (2015).

\bibitem{pointnet}
Qi, Charles R., et al. "Pointnet: Deep learning on point sets for 3d classification and segmentation". IEEE Conference on Computer Vision and Pattern Recognition (CVPR), 2017.

\bibitem{deepsdf}
Park, Jeong Joon, et al. "Deepsdf: Learning continuous signed distance functions for shape representation". IEEE Conference on Computer Vision and Pattern Recognition (CVPR), 2019.

\bibitem{occ_net}
Mescheder, Lars, et al. "Occupancy networks: Learning 3d reconstruction in function space". Proceedings of the IEEE Conference on Computer Vision and Pattern Recognition (CVPR), 2019.

\bibitem{pointr}
Yu, Xumin, et al. "Pointr: Diverse point cloud completion with geometry-aware transformers". International Conference on Computer Vision (ICCV), 2021.

\bibitem{marching_cubes}
William E. Lorensen and Harvey E. Cline. "Marching cubes:
A high resolution 3d surface construction algorithm". In SIGGRAPH. ACM, 1987.

\bibitem{bbox_sampling}
James J. Kuffner.
"Effective Sampling and Distance Metrics for 3D Rigid Body Path Planning". International Conference on Robotics and Automation (ICRA), 2004.


\bibitem{transformer}
Vaswani, Ashish, et al. "Attention is all you need." Advances in neural information processing systems, (NeurIPS), 2017.

\end{thebibliography}
\end{document}